\def\year{2020}\relax
\documentclass[letterpaper]{article} 
\usepackage{aaai22}  
\usepackage{times}  
\usepackage{helvet}  
\usepackage{courier}  
\usepackage[hyphens]{url}  
\usepackage{graphicx} 
\usepackage{natbib}  
\usepackage{caption} 
\DeclareCaptionStyle{ruled}{labelfont=normalfont,labelsep=colon,strut=off} 
\usepackage{algorithm}
\usepackage{algorithmic}
\usepackage{ dsfont }
\usepackage{multirow}
\usepackage{array}
\usepackage{amssymb}
\usepackage{microtype}
\usepackage{algorithm}
\usepackage{algorithmic}
\usepackage{graphicx}
\usepackage{ dsfont }
\usepackage{multirow}
\usepackage{array}
\usepackage{amssymb}
\usepackage{threeparttable}
\usepackage{changes}

\title{HAT4RD: Hierarchical Adversarial Training for Rumor Detection on\\ Social Media}
\author{
    Shiwen Ni, Jiawen Li and Hung-Yu Kao$^{\ast}$
}
\affiliations{
    \textsuperscript{\rm}Department of Computer Science and Information Engineering\\
    National Cheng Kung University\\
    Tainan, Taiwan \\
    \{P78083033, P78073012\}@gs.ncku.edu.tw, hykao@mail.ncku.edu.tw

}

\usepackage{bibentry}

\begin{document}

\maketitle

\begin{abstract}
With the development of social media, social communication has changed. While this facilitates people's communication and access to information, it also provides an ideal platform for spreading rumors. In normal or critical situations, rumors will affect people's judgment and even endanger social security. However, natural language is high-dimensional and sparse, and the same rumor may be expressed in hundreds of ways on social media. As such, the robustness and generalization of the current rumor detection model are put into question. We proposed a novel \textbf{h}ierarchical \textbf{a}dversarial \textbf{t}raining method for \textbf{r}umor \textbf{d}etection (HAT4RD) on social media. Specifically, HAT4RD is based on gradient ascent by adding adversarial perturbations to the embedding layers of post-level and event-level modules to deceive the detector. At the same time, the detector uses stochastic gradient descent to minimize the adversarial risk to learn a more robust model. In this way, the post-level and event-level sample spaces are enhanced, and we have verified the robustness of our model under a variety of adversarial attacks. Moreover, visual experiments indicate that the proposed model drifts into an area with a flat loss landscape, leading to better generalization. We evaluate our proposed method on three public rumors datasets from two commonly used social platforms (Twitter and Weibo). Experiment results demonstrate that our model achieves better results than state-of-the-art methods.
\end{abstract}

\section{Introduction}
Today, social media is a popular news source for many people. However, without automatic rumor detection systems, social media can be a breeding ground for rumors. Rumors can seriously affect people's lives \cite{ni2021mvan}. For instance, during the early outbreak of the current COVID-19 pandemic, rumors about a national lockdown in the United States fueled panic buying in groceries and toilet papers, disrupting the supply chain, exacerbating the demand-supply gap and worsening the issue of food insecurity among the socioeconomically disadvantaged and other vulnerable populations \cite{tasnim2020impact}. Setting up automatic rumor detection is therefore essential.

Automatic rumor detection is extremely challenging, and the greatest difficulty lies in spotting camouflaged rumors. As the saying goes, “\emph{Rumour has a hundred mouths}”, these words indicate that the ways rumors are expressed constantly change as they spread. Some malicious rumormongers may deliberately modify rumor text information to escape manual detection. Variability and disguise are the main characteristics of rumors, which means that a robust automatic rumor detection model is necessary.

\begin{figure*}[h]
	\centering
	\includegraphics[width=0.999\linewidth]{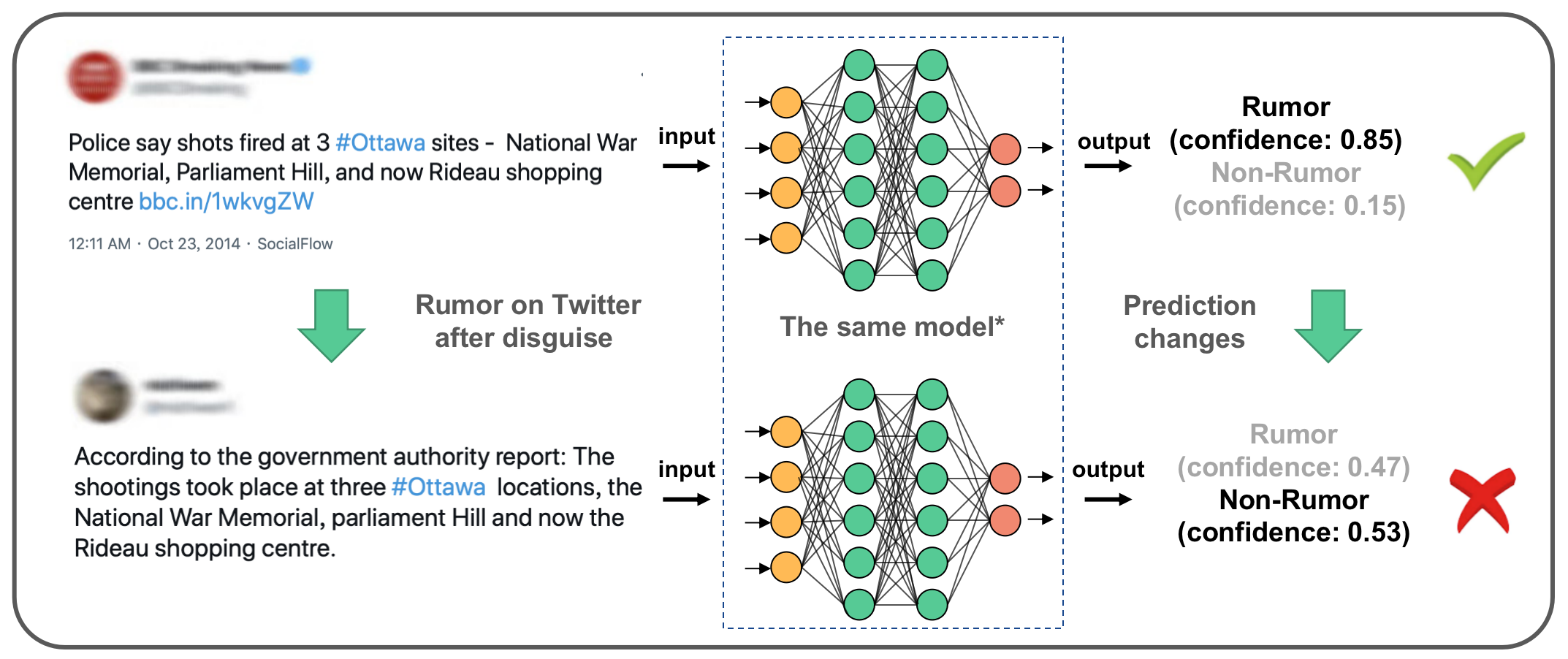}
	\caption{
		A well-trained BERT-base\protect\footnotemark[2] model for rumor detection. The rumor detection model label a rumor as a non-rumor when the words in rumor infinitesimal change but the meaning keep the same.
	}
	\label{f1}
\end{figure*}

Unfortunately, most current rumor detection models are not robust enough to spot the various changes and disguises during the rumor propagation process. As shown in Figure \ref{f1}, we simulated the constantly changing process of rumors during their propagation and found that the general deep learning model was too sensitive to sentence changes and disguise. A BERT-base \cite{devlin2019bert} model trained on the rumor dataset PHEME \cite{kochkina2018pheme} has a prediction confidence of 0.85 for the rumor \emph{"Police say shots fired at 3 \#ottawa sites National War Memorial, Parliament Hill, and now Rideau shopping centre"}, but when the input is changed to \emph{" According to the government authority report: The shootings took place at three \#ottawa locations the National War Memorial parliament Hill and now the Rideau shopping centre”}, model’s prediction confidence decreased from 0.85 to 0.47. However, the main meaning and label of the input rumor text did not change, but the model predicted incorrectly. This result shows that robustness and generalization of a traditional rumor detection model are poor, and the changes of a few words while the meaning of the sentence keeps the same may cause significant changes in the prediction results. 

\footnotetext[2]{The BERT-base code is completed by Hugging Face (https://github.com/huggingface/transformers).}

To alleviate that problem, we designed a novel rumor detection model called HAT4RD to enhance the generalization ability and robustness of an automatic rumor detection model. Our model detected rumors base on an event, which includes a source post and a certain number of replies. To make full use of the tweet object information and obtain a high-level representation, we took a hierarchical architecture as the skeleton of our model. To enhance robustness of our model, adversarial training is included in our model. Using more adversarial data to train the model can enhance the robustness and generalization of the model. However, natural language text space is sparse, and it is impossible to exhaust all possible changes manually to train a robust model. Thus, we perturb the sample space of post-level and event-level respectively to comprehensively improve the robustness of the model against changes in the text.
The main contributions of this paper can be summarized as follows: 

\begin{itemize}
	
	\item We first propose a hierarchical adversarial training method that encourages the model to provide robust predictions under the perturbed post-level and event-level embedding spaces.
	
	\item We evaluated the proposed model HAT4RD on three real-world datasets. The experimental results demonstrate that our model outperforms state-of-the-art models.
	
	\item We prove through experiments that the proposed hierarchical adversarial training method can enhance the robustness and generalization of the model and prevent the model from being deceived by disguised rumors.
	
\end{itemize}

\section{Related Work}
\subsection{Rumor Detection}
With the development of artificial intelligence, existing automated rumor detection methods are mainly based on deep neural networks. \citet{ma2016detecting} were the first to use a deep learning network, an RNN (Recurrent Neural Network)-based model, for automatic misinformation detection. \citet{chen2018call,yu2019attention} proposed an attention mechanism into an RNN or CNN (Convolutional Neural Network) model to process a certain number of sequential posts for debunking rumors. \citet{ajao2018fake} proposed a framework combining CNN and LSTM (Long Short Term Memory) to classify rumors. \citet{shu2019defend} delved into an explainable rumor detection model by using both news content and user comments. \citet{guo2018rumor,sujana2020rumor} detected rumors by creating a hierarchical neural network to obtain higher-level textual information representations. \citet{yang2018ti} proposed a rumor detection model that can handle both text and images. \citet{ruchansky2017csi} analyzed articles and extracted user characteristics to debunk rumors. \citet{ma2018rumor} constructed a recursive neural network to handle conversational structure. Their model was presented as a bottom-up and top-down propagation tree-structured neural network. \citet{li2020exploiting,li2020birds} used a variable-structure graph neural network to simulate rumor propagation and obtain more precise information representations in the rumor detection task. \citet{ni2021mvan} used multi-view attention networks to simultaneously capture clue words in the rumor text and suspicious users in the propagation structure. \citet{gumaei2022effective} proposed an extreme gradient boosting (XGBoost) classifier for rumor detection of Arabic tweets. \citet{li2021meet} combined objective facts and subjective views for an evidence-based rumor detection. No rumor detection model currently takes adversarial robustness into account. 
\subsection{Adversarial Training}
Adversarial training is an important method to enhance the robustness of neural networks. \citet{szegedy2014intriguing} first proposed the theory of adversarial training by adding small generated perturbations on input images. The perturbed image pixels were later named as adversarial examples. \citet{goodfellow2014explaining} proposed a fast adversarial example generation approach to try to obtain the perturbation value that maximizes adversarial loss. \citet{jia2017adversarial} were the first to adopt adversarial example generation for natural language processing tasks. \citet{zhao2018generating} found that when adopting the gradient-based adversarial training method on natural language processing tasks, the generated adversarial examples were invalid characters or word sequences. \citet{gong2018adversarial} utilized word vectors as the input for deep learning models, but this also generated words that could not be matched with any words in the word embedding space. \citet{ni2021dropattack} proposed a random masked weight adversarial training method to improve generalization of neural networks. However, so far there is no adversarial training method designed for rumor-specific hierarchical structures.

\section{Problem Definition}
Original rumors on social media are generally composed of a limited number of words and a few emojis. However, limited text information alone cannot accurately predict rumors. We, therefore, treat the original post and its reply posts together as an event for rumor detection. A whole event as the final decision-making unit contains a wealth of internal logic and user stance information. And the proposed hierarchical structure model starts with word embedding, forms post-embedding, event embedding, and finally predicts whether the event is a rumor through a fully connected layer.

Multiple events in the dataset are defined as $D=\{E_1,E_2,...,E_{|e|}\}$. An event consists of a source post and several reply posts, $E_j=\{P_s,P_1,P_2,...,P_{|p|}\}$. It should be noted that different events are composed of different numbers of posts, and a post is composed of different words, meaning our model needs to be able to process variable-length sequence information with a hierarchical structure. We consider the rumor detection task a binary classification problem. The event-level classifier can perform learning via labeled event data, that is, $E_j=\{P_s,P_1,P_2,...,P_{|p|}\}\rightarrow y_j$ . In addition, because an event contains multiple posts, we make the posts within the same event share labels. The post-level classifier $P_n=\{x_1,x_2,...,x_{|x|}\}\rightarrow y_n$ can therefore be established, and all the data will be predicted as the two labels: rumor or non-rumor.

\section{The Proposed Model HAT-RD}
\begin{figure}[h]
	\centering
	\includegraphics[width=0.95\linewidth]{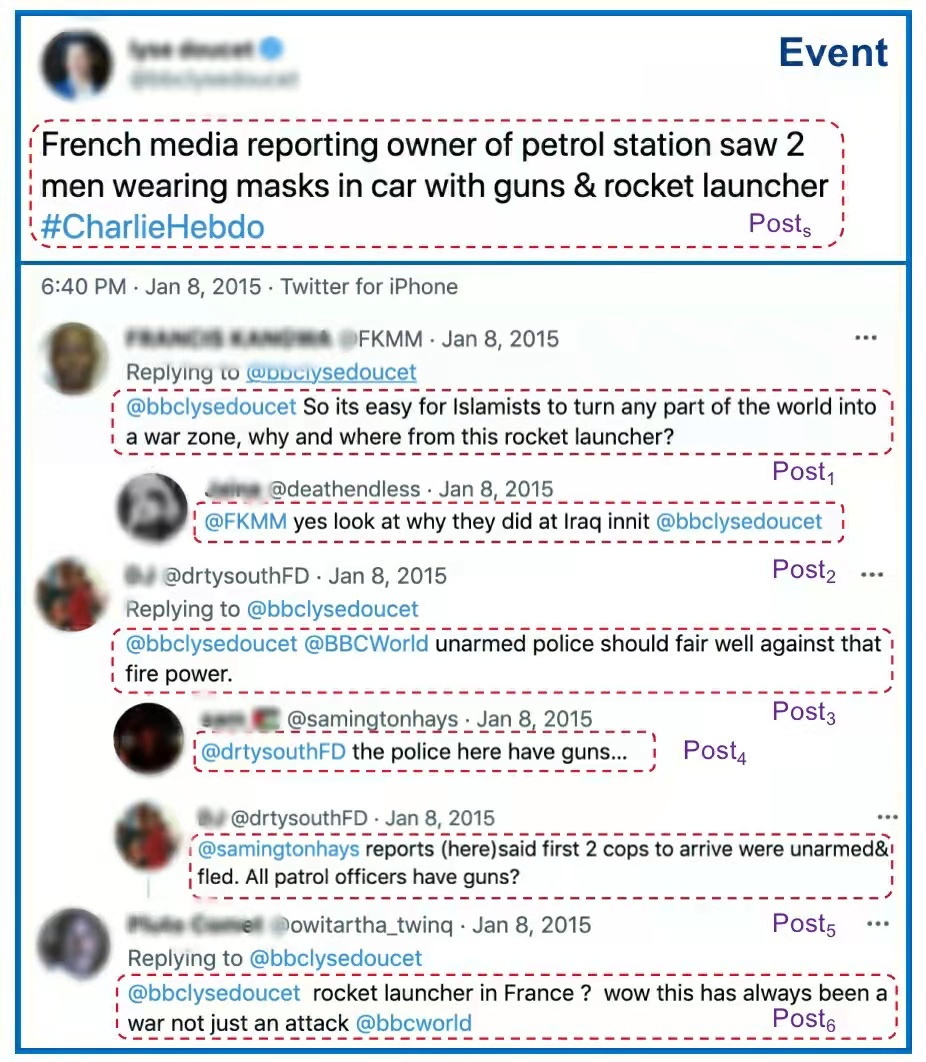}
	\caption{Posts and event on social media (hierarchical structure of post-level and event-level).
	}
	\label{f2}
\end{figure}

\begin{figure*}[t]
	\centering
	\includegraphics[width=0.99\linewidth]{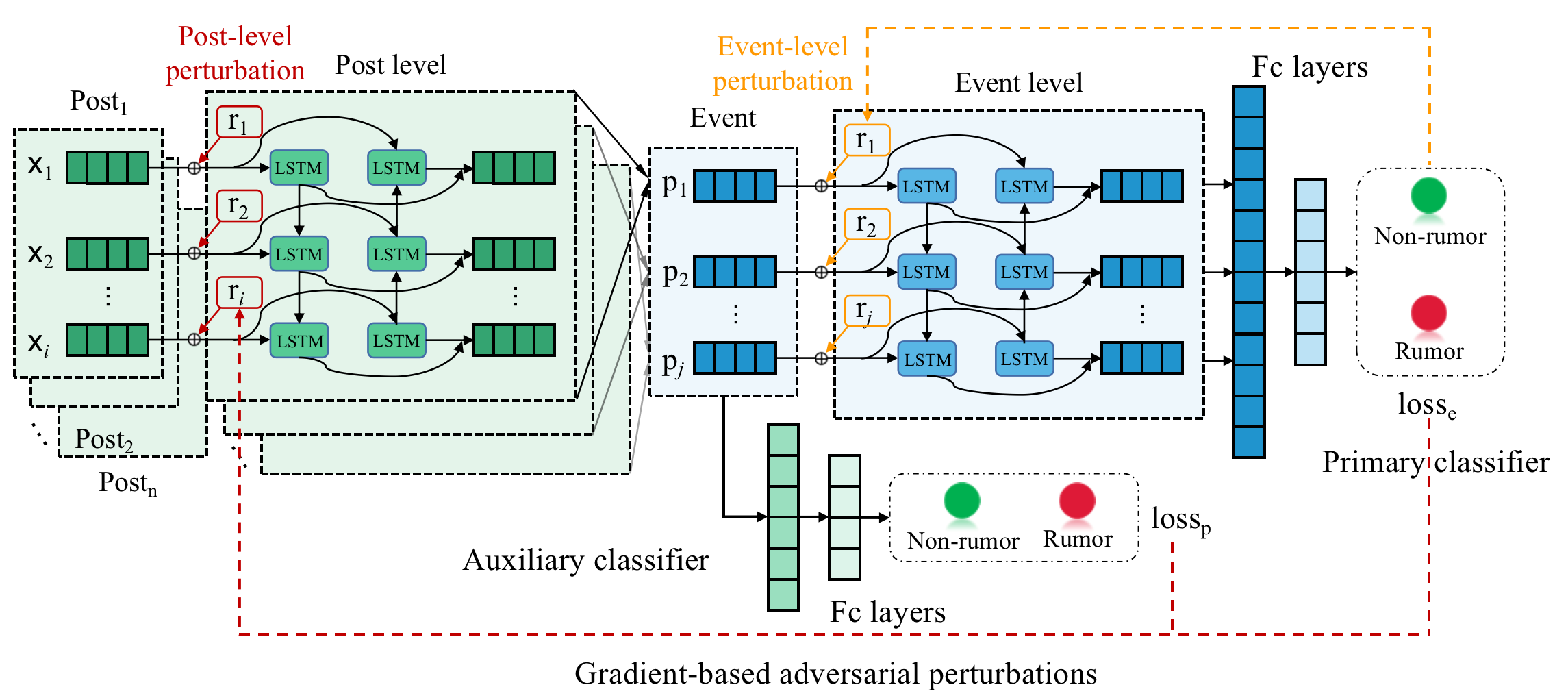}
\caption{The architecture of the proposed model HAT4RD.
	}
	\label{f2}
\end{figure*}

Rumors in social media have a hierarchical structure of post-level and event-level. Figure 2 shows a real-world rumor on Twitter. In response to this special data structure, we built the HAT-RD model based on the hierarchical BiLSTM, which can be divided into post-level modules and event-level modules, as shown in Figure 3. Hierarchical Adversarial Training (HAT) is a novel adversarial training method based on the hierarchical structure model. 
Taking the text of all posts under the event as input, we calculated the embedding of each word to obtain the input of post-level BiLSTM first. The formula is as follows:
\begin{equation}\label{key}
I_p=\{x_1,x_2,...,x_n\}
\end{equation}
where $I_p$ is the input of post-level BiLSTM, and all the vectors with the posts as the unit pass through the post-level BiLSTM layer in proper order. For each time point $t$, the formula is as follows:
\begin{equation}\label{key}
h^p_{t}={\rm BiLSTM_p}(x_i,h^p_{t-1})
\end{equation}

The cell state $h^p_{t}$ of the uppermost ${\rm LSTMp}$ at the last time point is used as the result of the post encoding. Due to the use of the bidirectional structure, the final state of both directions is joint, and an event can be represented by a matrix in which each column is a vector representing a post. The formula is as follows:
\begin{equation}\label{key}
O_p =[h^p_{s},h^p_{1},h^p_{2},...,h^p_{|p|}]
\end{equation}
where $h^p_{s}$ is the result of the post-level BiLSTM, that is, the embedding of the source post. $h^p_{i}$ is the embedding of a reply post, $O_p$ the output of post-level BiLSTM, and $I_e$ the input of event-level BiLSTM. The formula is as follows:
\begin{equation}\label{key}
I_e = O_p =[h^p_{s},h^p_{1},h^p_{2},...,h^p_{|p|}]
\end{equation}

For the next module, the event-level BiLSTM encoding process is similar to post-level BiLSTM. The difference can be seen in the input data unit; post-level BiLSTM uses a post vector composed of word vectors, while event-level BiLSTM uses an event vector composed of post vectors. The formula is as follows:
\begin{equation}\label{key}
h^e_{t}={\rm BiLSTM_e}(h^p_{t},h^e_{t-1})
\end{equation}

In the rumor detection classification task, the state $h^e_{t}$ of the event-level BiLSTM, the last layer at the last time point can be understood as a comprehensive representation of all posts.

Based on the principle of multi-task learning, rumor post classification and rumor event classification are highly related, and the parameters of the post-level module are shared in the two tasks. A post-level auxiliary classifier and an event-level primary classifier were therefore included in the hierarchical model. The post-level auxiliary classifier is mainly for accelerating training and preventing "vanishing gradient". Two classifiers were used to obtain post-level prediction results and event-level prediction results. The formula is as follows:
\begin{equation}\label{key}
\hat{y}_p = {\rm softmax}(W_p\cdot h^p_{t}+b_p)
\end{equation}
\begin{equation}\label{key}
\hat{y}_e = {\rm softmax}(W_e\cdot h^e_{t}+b_e)
\end{equation}
where $\hat{y_p}$ and $\hat{y_e}$ are the post and event classification results, respectively, $W_p$ and $W_e$ are the weights of the fully connected layers, and $b_p$ and $b_e$ are the biases. 
The goal of each training process is to minimize the standard deviation between the predicted and output values using the following loss function: 
\begin{equation}\label{key}
L_p = -ylog(\hat{y}_{p_r}-(1-y_p)log(1-\hat{y}_{p_n}))
\end{equation}
\begin{equation}\label{key}
L_e = -ylog(\hat{y}_{e_r}-(1-y_e)log(1-\hat{y}_{e_n}))
\end{equation}
\begin{equation}\label{key}
L_{t}=\alpha L_p + (1-\alpha)L_e
\end{equation}
where $L_p$ and $L_e$ are the post-level loss and event-level loss, respectively. $\alpha$ is the loss coefficient weight to control $L_p$ and $L_e$. $L_{t}$ is the total loss of the entire rumor detection model used to update the parameters. $y$ is the real label; $\hat{y}_{r}$ and $\hat{y}_{n}$ are the two labels predicted by the model -rumor and non-rumor. The gradient of the model was calculated according to Loss $L_{total}$. The formula is as follows:
\begin{equation}\label{key}
g=\nabla _{\theta}L_{t}(\theta,x,y)
\end{equation}
\begin{algorithm}[tb]
	\caption{Hierarchical adversarial training algorithm}
	\label{a1}
	\textbf{Input}: Training samples $\mathcal{X}$, perturbation coefficient $\epsilon_p$ and $\epsilon_e$, Loss coefficient weight $\alpha$,  Learning rate $\tau$ \\
	\textbf{Parameter}: $\theta$
	\begin{algorithmic}[1] 
		\FOR{${\rm epoch} = 1 \ldots N_{ep}$}
		\FOR{$(x,y)\in \mathcal{X}$}
		\STATE Forward-propagation calculation Loss:
		\STATE \hspace{2em}$L_p \leftarrow -ylog(\hat{y}_{p_r}-(1-y_p)log(1-\hat{y}_{p_n}))$
		\STATE	\hspace{2em}$L_e \leftarrow -ylog(\hat{y}_{e_r}-(1-y_e)log(1-\hat{y}_{e_n}))$
		\STATE	\hspace{2em}$L_{t}\leftarrow\alpha L_p + (1-\alpha)L_e$
		\STATE Backward-propagation calculation gradient:
		\STATE \hspace{2em}$g_p \leftarrow \nabla _{x_p}L_{t}(\theta,x_p,(y_p,y_e))$ 
		\STATE  \hspace{2em}$g_e \leftarrow\nabla _{x_e}L_{e}(\theta,x_e,y_e)$
		\STATE Compute perturbation:
		\STATE \hspace{2em}$r_p\leftarrow\epsilon_p\cdot g_p/||g_p||_2$
		\STATE \hspace{2em}$r_e\leftarrow\epsilon_e\cdot g_e/||g_e||_2$
		\STATE Forward-Backward-propagation calculation adversarial gradient:
		\STATE \hspace{2em}$g^p_{adv}\leftarrow\nabla_\theta L^p_{t_{adv}}(\theta,x_p+r_p,(y_p,y_e))$
		\STATE \hspace{2em}$g^e_{adv}\leftarrow\nabla _\theta L^e_{e_{adv}}(\theta,x_e+r_e,y_e)$
		\STATE Update parameter:
		\STATE \hspace{2em} $\theta \leftarrow \theta -\tau(g+g^p_{adv}+g^e_{adv})$
		\ENDFOR
		\ENDFOR
		\STATE \textbf{Output}: $\theta$
	\end{algorithmic}
\end{algorithm}
\subsection{Hierarchical Adversarial Training}
The above is a forward propagation under standard training of the model. We needed to make the model perform hierarchical adversarial training. The overall hierarchical adversarial training procedure is shown in Algorithm \ref{a1}. This adversarial optimization process was expressed with the following Min-Max formula:
\begin{equation}\label{}
\mathop{{\rm min}}\limits_{\theta} \mathds{E}_{(x,y)\sim D}\{\mathop{{\rm max}}\limits_{r_{p},r_{e}\in S}[L_{t}(\theta,x_p+r_p,(y_p,y_e)) +L_{e}(\theta,x_e+r_e,y_e) ]\}
\end{equation}
where $r_p$ and $r_e$ are the perturbations of the post-level input $x_p$ and event-level input $x_e$ under maximization of the internal risk. We respectively estimated these values by linearizing $\nabla _{x_p}L_{t}(\theta,x_p,(y_p,y_e))$ and  $\nabla _{x_e}L_{e}(\theta,x_e,y_e)$ around $x_p$ and $x_e$. Using the linear approximation  $\nabla _{x_p}L_{t}(\theta,x_p,(y_p,y_e))$ and $\nabla _{x_e}L_{e}(\theta,x_e,y_e)$  in equation (13), (14) and the L2 norm constraint, the resulting adversarial perturbations are:
\begin{equation}\label{key}
r_p=\epsilon_p\cdot \frac{ \nabla _{x_p}L_{t}(\theta,x_p,(y_p,y_e))}{||\nabla _{x_p}L_{t}(\theta,x_p,(y_p,y_e))||_2}
\end{equation}
\begin{equation}\label{key}
r_e=\epsilon_e\cdot \frac{ \nabla _{x_e}L_{e}(\theta,x_e,y_e)}{||\nabla _{x_e}L_{e}(\theta,x_e,y_e)||_2}
\end{equation}
where $\epsilon_p$ and $\epsilon_e$ are the perturbation coefficients. Note that the value of the perturbation $r_p$ is calculated based on the back-propagation of the total Loss  instead $L_t$ of $L_p$, because the addition of the perturbation $r_p$ makes $L_p$ and Le increase at the same time.
\subsubsection{Post-Level Adversarial Training}
After a normal forward and backward propagation,  and  were calculated according to the gradient. Using post-level adversarial training, we added word-level perturbation to the word vector to obtain the input of post-level BiLSTM, and the formula is as follows:
\begin{equation}\label{key}
I_{p_{adv}}=\{x_1+r^p_1,x_2+r^p_2,...,x_n+r^p_n\}
\end{equation}
where $I_{p_{adv}}$ is the adversarial input of post-level BiLSTM, and $r^p_n$ is the post-level perturbation added to the word vector $x_n$. All the vectors with the posts as the unit then pass through the post-level BiLSTM layer in proper order. For each time point $t$, the formula is as follows:
\begin{equation}\label{key}
h^{p_{adv}}_t={\rm BiLSTM_p}(x_i+r^p_i,h^{p_{adv}}_{t-1})
\end{equation}
The adversarial cell state $h^{p_{adv}}_t$ of the uppermost ${\rm LSTM_p}$ at the last time point is used as the result of the post encoding. Due to the use of the bidirectional structure, the final state of both directions is joint, and an event can be represented by a matrix in which each column is a vector representing a post. The formula is as follows:
time point $t$, the formula is as follows:
\begin{equation}\label{key}
O_{p_{adv}}=[h^{p_{adv}}_s,h^{p_{adv}}_1,h^{p_{adv}}_2,...,h^{p_{adv}}_{|p|}]
\end{equation}
where $h^{p_{adv}}_s$ is the adversarial result of the post-level BiLSTM, that is, the embedding of the source post. $h^{p_{adv}}_i$ is the adversarial embedding of the reply post, and $O_{p_{adv}}$ is the adversarial output of post-level BiLSTM and input of event-level BiLSTM. The formula is as follows:
\begin{equation}\label{key}
h^{e_{adv}}_t={\rm BiLSTM_p}(h^{p_{adv}}_t+r^e_t,h^{e_{adv}}_{t-1})
\end{equation}
Finally, $h^e_t$ was replaced with $h^{e_{adv}}_t$ and the adversarial loss $L^p_{p_{adv}}$, $L^p_{e_{adv}}$ and $L^p_{t_{adv}}$ of post-level perturbation can be calculated using equations (6)-(9). The post-level adversarial gradient $g^p_{adv}$ is calculated based on the result of  backpropagation. The formula is as follows:
\begin{equation}\label{key}
g^p_{adv}=\nabla_\theta L^p_{t_{adv}}(\theta,x_p+r_p,(y_p,y_e))
\end{equation}
\subsubsection{Event-Level Adversarial Training}
We next performed event-level adversarial training and repeated the process of equations (1), (2) and (3) to obtain the posts vector. Event-level perturbation was then added to the post vector to obtain the adversarial input of event-level BiLSTM, and the formula is as follows:
\begin{equation}\label{key}
I_{e_{adv}}=\{h^p_s+r^p_s,h^p_1+r^p_1,h^p_2+r^p_2,...,h^p_{|p|}+r^p_{|p|}\}
\end{equation}
In the same way, input $I_{e_{adv}}$ into the event-level BiLSTM to get the final event representation vector $h^{e_{adv}}_t$, replace $h^e_t$ with $h^{e_{adv}}_t$ and calculate the adversarial loss $L^e_{e_{adv}}$ of event-level perturbation through equations (6)-(9). Finally, the post-level adversarial gradient $g^e_{adv}$ is calculated based on backpropagation. The formula is as follows:
\begin{equation}\label{key}
g^e_{adv}=\nabla _\theta L^e_{e_{adv}}(\theta,x_e+r_e,y_e)
\end{equation}
Finally, the gradient is calculated by the standard training; the gradient calculated by the post-level adversarial training and the gradient calculated by the event-level adversarial training were used to update the model parameters. The parameter update process is expressed as:
\begin{equation}\label{key}
\theta \leftarrow \theta -\tau(g+g^p_{adv}+g^e_{adv})
\end{equation}
where $\tau$ is the learning rate.

\section{Experiments}
\subsection{Datasets}
Three well-known public rumor datasets, PHEME 2017, PHEME 2018 \citep{kochkina2018pheme} and WEIBO \citep{ma2016detecting} , are used to evaluate our method HAT4RD. Among them, the original data of PHEME 2017 and PHEME 2018 are from the Twitter social platform, and the language is English; the original data of WEIBO is from the Sina Weibo social platform, and the language is Simplified Chinese. In these three datasets, each event is composed of a source post and several reply posts, The statistical details of these three datasets are shown in Table \ref{t1}. 
"Users" represents the number of users in the datasets; "Posts" represents the number of posts in the datasets; "Event" represents the number of events in the datasets (that is, the number of source posts); Avg words/post" represents the average number of words contained in a post; "Avg posts/event" represents the average number of posts contained in an event; "Rumor" represents the number of rumors in the datasets; "Non-rumor" represents the number of non-rumors in the datasets; "Balance degree" represents the percentage of rumors in the datasets.

\begin{table*}[h]
	\centering
	\renewcommand\arraystretch{1.1}
	\caption{The statistics of datasets used in this paper.}
	\setlength\tabcolsep{14 pt} 
	\begin{tabular}{lrrr}
		\hline \hline
		\textbf{Statistic} & \multicolumn{1}{c}{\textbf{PHEME 2017}} & \multicolumn{1}{c}{\textbf{PHEME 2018}} & \multicolumn{1}{c}{\textbf{WEIBO}}\\ \hline
		\textbf{Users} & 49,345 & 50.593&2,746,818 \\
		\textbf{Posts} & 103,212 & 105,354&3,805,656 \\
		\textbf{Events} & 5,802 & 6,425 &466,4\\
		\textbf{Avg words/post} & 13.6 & 13.6&  23.2\\
		\textbf{Avg posts/event} & 17.8 & 16.3 &816.0\\
		\textbf{Rumor} & 197,2 & 240,2&231,3 \\
		\textbf{Non-rumor} & 383,0 & 402,3&235,1 \\
		\textbf{Balance degree} & 34.00\% & 37.40\%& 49.59\% \\ \hline \hline
	\end{tabular}
	\label{t1}
\end{table*}

\subsection{Experimental Settings}
Following the work of \citep{li2021meet}, the datasets were split for our experiment: 80\% for training, 10\% for validation, and 10\% for testing.  We trained all the models by employing the derivative of the loss function through backpropagation and used the Adam optimizer \citep{kingma2014adam} to update the parameters. From post text to its embedding, we use Glove's \citep{pennington2014glove} pre-trained 300-dim word vector. For the hyperparameters, the maximum value of vocabulary is 80000; the batch size is 64, the dropout rate is 5,. the BiLSTM hidden size unit is 512, the loss coefficient weight $\alpha$ is 0.1, the learning rate is 0.0001, and the perturbation coefficient $\epsilon_p$ and $\epsilon_e$ are 1.0 and 0.3. Our proposed model was finally trained for 100 epochs with early stopping. In addition, all experiments are run under the following hardware environment: CPU: Intel(R) Core(TM) i7-8700 CPU@3.20GHz, GPU: GeForce RTX 2080, 10G.

\subsection{Performance Comparison}
Our HAT4RD model was compared with other well-known rumor detection models to evaluate our model's rumor debunking performance. 
\begin{itemize}
	\item SVM-BOW: a rumor detection naive baseline, which is an SVM that uses bag-of-words for word representation \citep{ma2018rumor}. 
	
	\item TextCNN: a rumor detection naive baseline based on deep convolutional neural networks. \citep{chen2017ikm}.

	\item BiLSTM: a rnn-based bidirectional model that detects rumor by considering the bidirectional information \citep{augenstein2016stance}.
	
	\item BERT: a well-known pre-trained language model. We fine-tuned a BERT-base to detect rumors \citep{devlin2019bert}.
	
	\item CSI: a state-of-the-art model detecting rumor by scoring users based on their behavior \citep{ruchansky2017csi}.
	
	\item CRNN: a hybrid model that combines recurrent neural network and convolutional neural network to detect rumors \citep{liu2018early}.
	
	\item RDM: a rumor detection model that integrates reinforcement learning and deep learning for early rumor detection \citep{zhou2019early}.
	
	\item CSRD: a rumor detection model that classifies rumors by simulating comments' conversation structure using GraphSAGE \citep{li2020exploiting}.
	
	\item EHCS-Con: a model exploited the user’s homogeneity by using the node2vec mechanism encoding user’s follow-followers relationship for rumor detection \citep{li2020birds}.
	
	\item LOSIRD: a state-of-the-art rumor detection model that leverages objective facts and subjective views for interpretable rumor detection \citep{li2021meet}.
	
\end{itemize}
\subsection{Main Experiment Results}
The results of different rumor detection models are compared in Table \ref{t2}; the HAT4RD clearly performs the best in terms of rumor detection compared to the other methods based on the three datasets with 92.5\% accuracy on PHEME 2017 , 93.7\% on PHEME 2018 and 94.8\% on WEIBO. In addition, the precision, recall, and F1 are all higher than 91\% in the HAT4RD model. Our HAT4RD improves the F1 of the SOTA model by about 1.5\% on the dataset WEIBO. These results demonstrate the effectiveness of the hierarchical structure model and hierarchical adversarial training in rumor detection. However, the SVM-BOW result is poor because the traditional statistical machine learning method could not handle this complicated task. The results of the CNN, BiLSTM, BERT and RDM models are poorer than ours due to their insufficient information extraction capabilities. The models are based on post-processing information and cannot get a high-level representation from the hierarchy. Compared to other models, our HAT4RD model has a hierarchical structure and performs different levels of adversarial training. This enhances both post-level and event-level sample space and improves the robustness and generalization of the rumor detection model. 

\begin{table*}[h]
		\centering
	\begin{threeparttable}
		\centering
		\renewcommand\arraystretch{1.0}
		\setlength\tabcolsep{8 pt} 
		\caption{The results of different methods on three datasets. We report their average of 5 runs.}
		\label{t2}
		\begin{tabular}{l|c|c|c|c|c|c|c|c|c|c|c|c}
			\hline \hline
			\multirow{2}{*}{\textbf{Method}} & \multicolumn{4}{c|}{\textbf{PHEME 2017}} & \multicolumn{4}{c|}{\textbf{PHEME 2018}}&\multicolumn{4}{c}{\textbf{WEIBO}} \\ \cline{2-13}
			& \textbf{Acc} & \textbf{Pre} & \textbf{Rec} & \textbf{F1} & \textbf{Acc} & \textbf{Pre} & \textbf{Rec} & \textbf{F1} &  \textbf{Acc} & \textbf{Pre} & \textbf{Rec} & \textbf{F1} \\ \hline
			SVM-BOW & 66.9 & 53.5 & 52.4 & 51.9 & 68.8 & 51.8 & 51.2 & 50.4 & 72.3 & 63.5 & 67.4 & 65.6 \\
			TextCNN & 78.7& 73.7& 70.2& 71.0& 79.4& 73.2& 67.3& 68.6& 84.2& 73.3& 77.9& 75.5 \\
			BiLSTM  & 79.5& 76.3& 69.1& 70.6& 79.6& 72.7& 67.7& 68.9& 85.7& 83.1& 89.6& 86.4\\
			BERT  & 86.5& 85.9& 85.1& 85.5& 84.4& 83.4& 83.5& 83.5& 90.7& 89.4& 89.7& 89.5\\
			CSI & 85.7& 84.3& 85.9& 85.1& 85.1& 83.6& 85.5& 84.5& 91.4& 90.4& 90.7& 90.5\\
			CRNN & 85.5& 84.6& 85.4& 85.0& 86.2& 85.7& 85.6& 85.6& 91.1& 90.2& 91.8&91.0 \\
			RDM & 87.3& 81.7& 82.3& 82.0& 85.8& 84.7& 85.9& 85.2& 92.7& 91.6& 93.7&92.6 \\
			CSRD & 90.0 & 89.3 & 86.9 & 88.1 & 91.9 & 89.2 & 92.3 & 90.7& 92.4& 91.5& 91.7 &91.6  \\
			EHCS-Con & 91.2 & 90.5 & 90.5 & 90.5 & 92.3& 92.3 & 92.3 & 92.4 & 93.0& 92.2& 92.6 &92.4  \\
			LOSIRD \tnote{$\dagger$}  & 91.4 & 91.5   & 90.0  & 90.6  & 92.5  & 92.2   & 92.4  & 92.3 & 93.2 & 92.3 & 92.7  &92.5    \\ 
			\textbf{HAT4RD} \tnote{$\ddagger$} & \textbf{92.5} & \textbf{91.7} & \textbf{91.1} & \textbf{91.7} & \textbf{93.8} & \textbf{93.1} & \textbf{93.6} & \textbf{93.4}& \textbf{94.8} & \textbf{93.8} & \textbf{94.2} & \textbf{94.0} \\ \hline \hline
		\end{tabular}
		\begin{tablenotes}
			\footnotesize
			\item[$\dagger$] {The state-of-the-art model}; $^\ddagger$ {Our model. } 
		\end{tablenotes}
	\end{threeparttable}
\end{table*}

\subsection{Ablation Analysis}
\begin{figure}[h]
	\centering
	\includegraphics[width=0.95\linewidth]{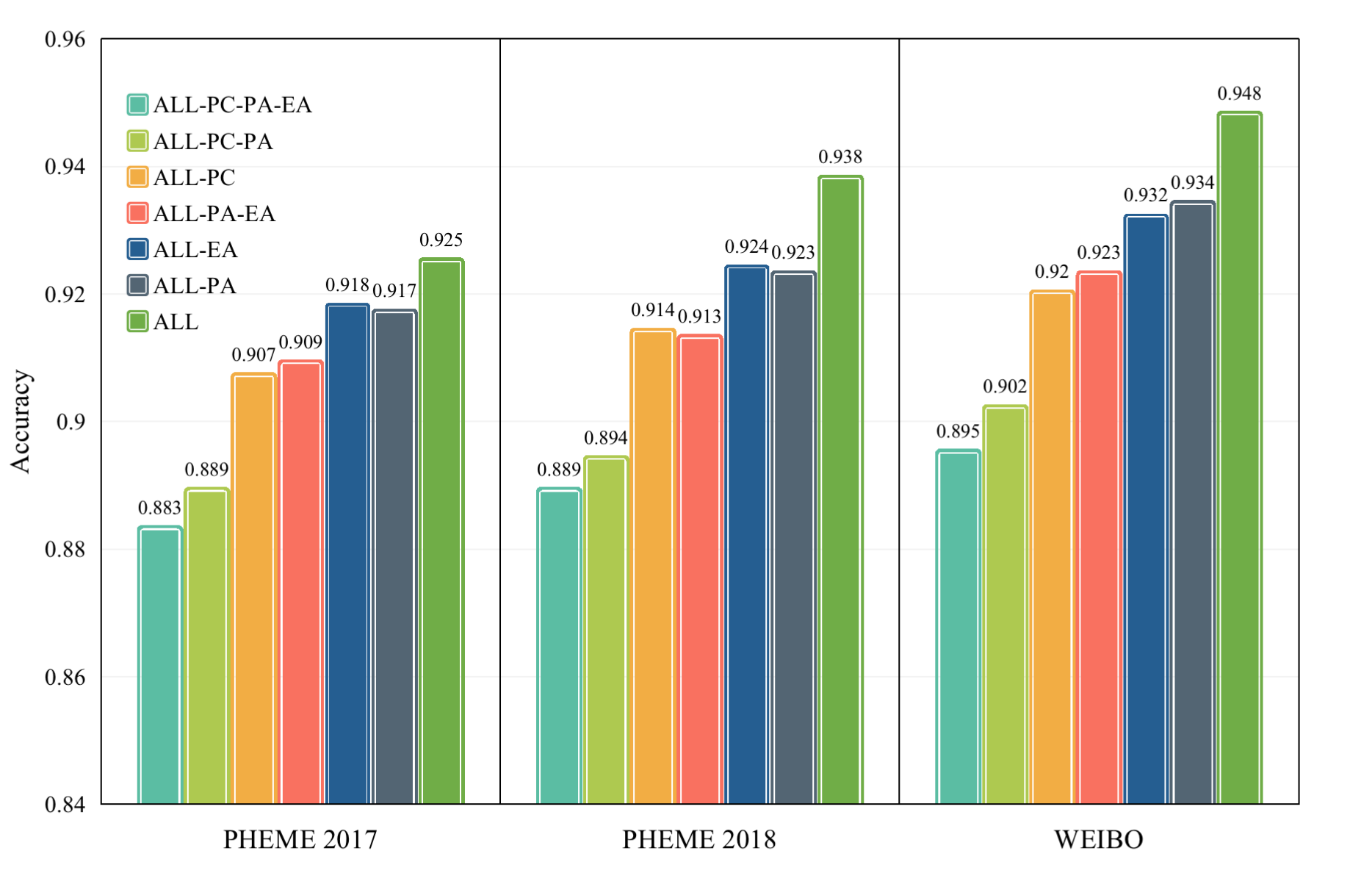}
	\caption{HAT4RD ablation analysis in accuracy.
	}
	\label{f4}
\end{figure}

To evaluate the effectiveness of every component of the proposedHAT4RD, we removed each one of them from the entire model for comparison. "ALL" denotes the entire model HAT4RD with all components, including post-level adversarial training (PA), event-level adversarial training (EA), the post-level auxiliary classifier (PC), and event-level primary classifier (EC). After the removing, we obtained the sub-models "-PA", "-EA", "-PC" and "-EC", respectively. "-PA-PC" means that both the post-level adversarial training and auxiliary classifier were removed. "-PA-EA" denotes the reduced HAT4RD without both post-level adversarial training and event-level adversarial training. The results are shown in Fig. \ref{f4} . 

It can be observed that every component plays a significant role in improving the performance of HAT4RD. HAT4RD outperforms ALL-PA and ALL-EA, which shows that the post-level adversarial training and event-level adversarial training are indeed helpful in rumor detection. Both ALL-PA and ALL-EA are better than ALL-PA-EA, which shows that hierarchical adversarial training is more efficient than single-level adversarial training. The performance of ALL-PC is lower than that of HAT4RD, proving that the post-level auxiliary classifier contributes to the learning and convergence of the model. 

\subsection{Early Rumor Detection}
\begin{figure*}[]
		\centering
		\includegraphics[width=1\linewidth]{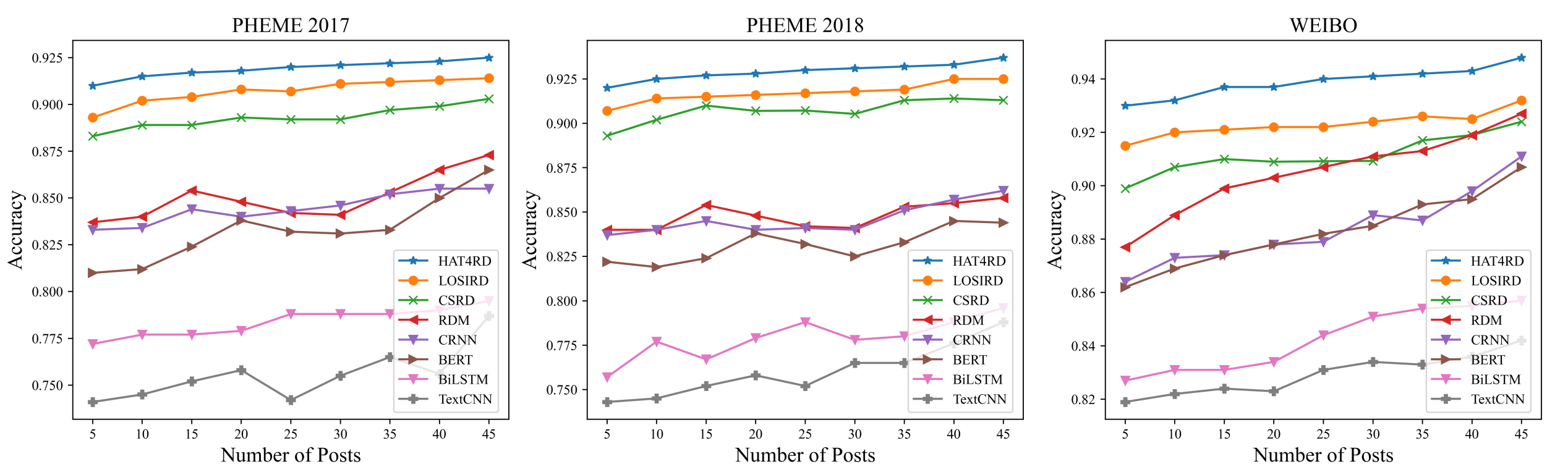}
		\caption{Early rumor detection accuracy.
		}
		\label{f5}
\end{figure*}

Our model's performance in early rumor detection was evaluated. To simulate the early stage rumor detection scenarios in the real world, 9 different size test sets from PHEME 2017, PHEME 2018 and WEIBO were created. Each test set contained a certain number of posts, ranging from 5 to 45. We found that HAT4RD could detect rumors with an approximate 91\% accuracy rate with only 5 posts as illustrated in Fig. \ref{f5}. Compared to the other models, our model uses hierarchical adversarial training and continuously generates optimal adversarial samples to join the training. It, therefore, has good generalization despite limited information.

\footnotetext[3]{https://github.com/thunlp/OpenAttack}
\subsection{Robustness Analysis}
We use \emph{OpenAttack}\footnotemark[3] \cite{zeng2020openattack} to conduct a variety of adversarial attacks on the models, and compare the robustness of various recent models. FSGM  draws from \citep{goodfellow2014explaining}, which is a gradient-based adversarial attack method. HotFlip \cite{ebrahimi2018hotflip} uses gradient-based word or character substitution to attack. PWWS \cite{ren2019generating} uses a greedy word substitution order determined by the word saliency and weighted by the classification probability. As shown in Table \ref{t3}, our model can maintain the minimum performance degradation under the three adversarial attacks compared to other baseline models. Especially for gradient-based attacks, the robustness of our model is very obvious. Under the attack of FSGM, the performance of our model only dropped by about 10\%. Under the attacks of HotFlip and PWWS, our model HAT4RD is also significantly more robust than other models.
\begin{table*}[h]
	\caption{Classification accuracy of models on the PHEME 2017, PHEME 2018 and WEIBO datasets and the perturbed datasets using different attacking methods.}
	\setlength{\tabcolsep}{1.3mm}{
		\begin{tabular}{l|cccc|cccc|cccc}
			\hline \hline
			\multirow{2}{*}{\textbf{Method}} & \multicolumn{4}{c|}{\textbf{PHEME 2017}} & \multicolumn{4}{c|}{\textbf{PHEME 2018}} & \multicolumn{4}{c}{\textbf{WEIBO}} \\  \cline{2-13}
			&Original & FSGM & HotFlip & PWWS& Original & FSGM & HotFlip & PWWS& Original & FSGM & HotFlip & PWWS \\ \hline 
			Bert & 0.865 & 0.432 & 0.426 & 0.425& 0.834 & 0.412 & 0.431 & 0.442 & 0.907 & 0.513 & 0.442 & 0.474 \\
			EHCS-Con & 0.912 & 0.547 & 0.576 & 0.531& 0.923 & 0.568 & 0.513 & 0.412 & 0.930 & 0.673 & 0.543 & 0.484 \\
			LOSIRD & 0.914 & 0.576 & 0.415 & 0.287 & 0.922 & 0.534 & 0.501 & 0.491 & 0.932 & 0.624 & 0.654 & 0.446\\
			HAT4RD & 0.925 & 0.846 & 0.786 & 0.534& 0.932 & 0.835 & 0.744 & 0.615 & 0.948 & 0.853 & 0.726 & 0.696\\ \hline \hline
	\end{tabular}} 
	\label{t3}
\end{table*}
\footnotetext[4]{https://github.com/tomgoldstein/loss-landscape}
\subsection{The Impact of Hierarchical Adversarial Training on Loss Landscape}
\begin{figure*}[h]
		\centering
		\includegraphics[width=0.99\linewidth]{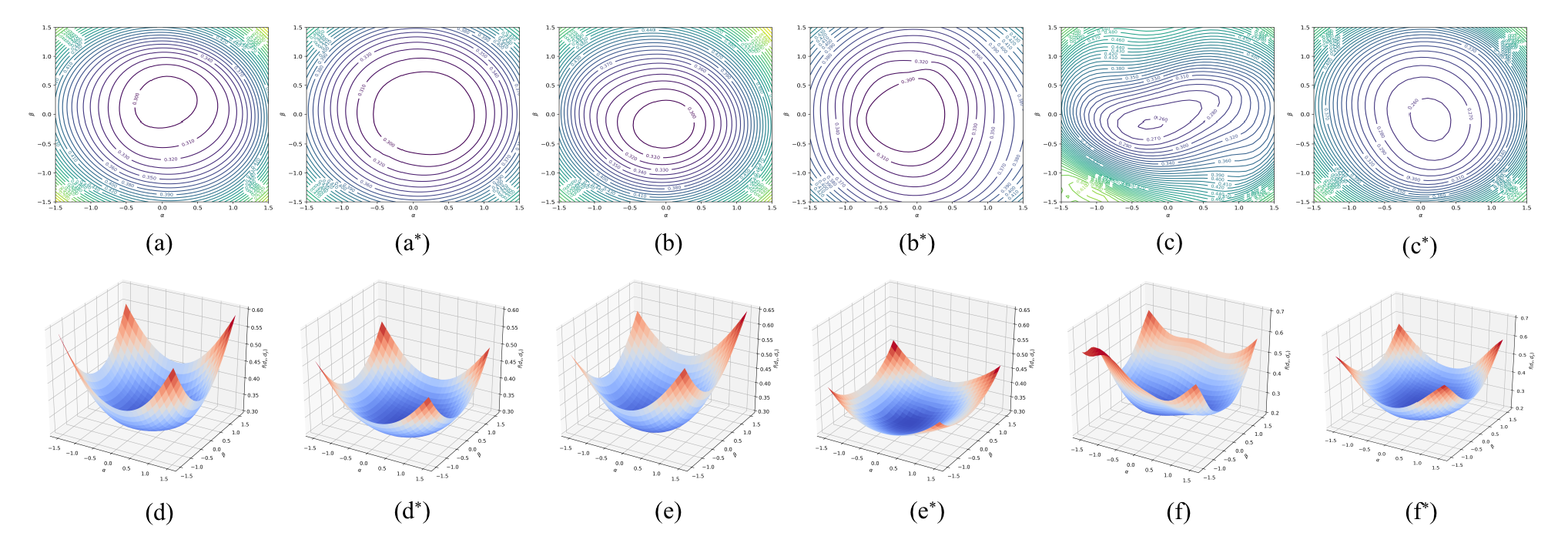}
		\caption{2D and 3D visualization of the minima of the Loss function selected by standard training (a, b, c, d, e, f) and hierarchical adversarial training (a*, b*, c*, d*, e*, f*) on PHEME 17 (a, a*, d, d*), PHEME 18 (b, b*, e, e*) and WEIBO (c, c*, f, f*) datasets.
		}
		\label{f6}
\end{figure*}

To further visually analyze the effectiveness of the hierarchical adversarial training method, we drew the high-dimensional non-convex loss function with a visualization method\footnotemark[4] proposed by \cite{li2018visualizing}. We visualize the loss landscapes around the minima of the empirical risk selected by standard and hierarchical adversarial training with the same model structure. The 2D and 3D views are plotted in Figure \ref{f6}. We defined two direction vectors, $d_x$ and $d_y$ with the same dimensions as $\theta$, drawn from a Gaussian distribution with zero mean and a scale of the same order of magnitude as the variance of layer weights. We then chose a center point $\theta^\ast$ and added a linear combination of $\alpha$ and $\beta$ to obtain a loss that is a function of the contribution of the two random direction vectors.

\begin{equation}\label{key}
f(d_x,d_y)=L(\theta^\ast+\alpha d_x+\beta d_y)
\end{equation}
The results show that the hierarchical adversarial training method indeed selects flatter loss landscapes by dynamically generating post-level perturbation and event-level perturbation.  Having a flatter Loss function indicates that the model is more robust in input features and can prevent the model from overfitting. Empirically, many studies have shown that a flatter loss landscape usually means better generalization \cite{hochreiter1997long,keskar2019large,ishida2020we}.
\section{Conclusion and Future Work}

Herein, we proposed a new hierarchical adversarial training for rumor detection that considers the camouflages and variability of rumors from an adversarial perspective. Dynamically generating perturbations on the post-level and event-level embedding vectors enhanced the model's robustness and generalization.  The evaluations of three real-world rumor detection datasets on social media show that our HAT4RD model outperforms state-of-the-art methods. Numerically, our proposed HAT4RD is 1.1\%, 1.1\%, and 1.5\% higher than the F1 of the state-of-the-art  model LOSIRD on three public rumor detection datasets, respectively. The early rumor detection performance of our model also outperforms other models. We point out the contribution of each part to model performance through ablation experiments. Moreover, visual experiments prove that the hierarchical adversarial training method we proposed can optimize the model for a flatter loss landscape. Our HAT4RD model is general and can be applied to data on any topic, as long as the data is posted on social media (e.g., Twitter, Weibo). The ability of our model depends on the training dataset. We only need to add the corresponding data to the model training to detect rumors of different topics.

Robustness and generalization are the focus of rumor detection. In the future, we can integrate features such as text and images for multi-modal adversarial training to further enhance the model. In addition, for the unique structure of posts and events, we think that graph neural networks will also be a good research direction, and graph neural networks can be combined with adversarial training to obtain graph adversarial training. We think this will be an interesting research direction. Because rumor data collection and labeling are complicated and time-consuming, the recently popular prompt learning based on pre-trained language models for few-shot rumor detection is also worth studying.

\section{Limitation}
Finally, our model currently has certain limitations. Since our model includes hierarchical adversarial training, the training time is longer than the general model. Moreover, although our hierarchical adversarial training improves the robustness of the model, our model still has a lot of room for improvement due to the diversity of rumors and the sparsity of natural language.

Robustness and generalization are the focus of rumor detection. In the future, we can integrate features such as text and images for multi-modal adversarial training to further enhance the model. 

\bibliography{aaai22}

\bigskip

\end{document}